\pdfoutput=1
\documentclass{article} 
\usepackage{iclr2027_conference,times}

\usepackage{hyperref}
\hypersetup{colorlinks=true, linkcolor=blue, citecolor=blue, urlcolor=blue,
            breaklinks=true}
\usepackage{url}
\usepackage{booktabs}
\usepackage{multirow}
\usepackage{graphicx}
\usepackage{xcolor}
\usepackage{amsmath,amssymb}

\newcommand{\ngym}{NeutronGym}
\newcommand{\mcbench}{McStasBench}
\newcommand{\instr}{\texttt{.instr}}

\title{NeutronGym: Physics-Graded Neutron\\
Instrument Design for LLM Agents}

\author{Lijie Ding$^{\ast}$ \& Changwoo Do \\
Neutron Scattering Division, Oak Ridge National Laboratory, Oak Ridge, TN 37831, USA \\
$^{\ast}$\texttt{dingl1@ornl.gov}}

\iclrfinalcopy
\begin{document}

\maketitle

\lhead{}

\begin{abstract}
Designing a scientific instrument tests whether language-model agents can do
physics rather than recall it, provided the grading cannot be argued with. We
introduce \ngym{}, to our knowledge the first executable environment for
neutron instrument design: agents build instruments through validating tools,
McStas ray-traces what they build, and a level-resolved ladder grades syntax,
runtime, structure and science with no LLM judge. Procedural families supply
unlimited instances of a fixed layout whose design parameters the agent must
set, with held-out parameter regimes; a curated slice, \mcbench{}, adds 16
tasks from published instruments behind memorization probes and a sandbox.
Seven models reproduce at most 7 of the 16, none retrieves a reference, and
none meets an improvement target. The environment also trains. Reinforcement
learning on its reward takes Qwen3-8B from 11\% to 77\% of held-out instances
of a family whose targets come from a hidden design (69\% at a second seed),
past an untrained Qwen3-32B, and the recipe holds, at one seed each, on three
further gated families. The analysis says what that gain is. Without the
ladder's partial credit it collapses by 60 points. From reward alone the
trained model reaches what a classical optimizer reaches, at the agent's
simulation budget, only when handed the closed-form physics (77\% against
81\%, a gap that does not separate at this size), while frontier models still
solve 98--99\%. Getting a trustworthy result meant failing four task designs
that no-model baselines could solve, and we release the probes that found
them.
\end{abstract}
\section{Introduction}
\label{sec:intro}

Designing a scientific instrument asks more of an agent than writing code that
runs: it must turn a specification into a physical layout, trade competing
constraints against each other, and produce a design whose quality is measured
by physics rather than by agreement with a reference answer. Neutron
scattering instruments are an unusually clean substrate for this. They are
designed in simulation long before construction, the community-standard
simulator McStas~\citep{lefmann1999mcstas,willendrup2020mcstas} describes an
instrument as a text file of components, a Monte Carlo ray trace returns
quantitative observables (intensity, divergence, wavelength distribution) in
seconds, and passive neutron optics obey a conservation law that gives the
deliverable intensity a computable ceiling needing no reference answer.

Turning such a simulator into an evaluation and training substrate raises three
problems that are easy to underestimate. \emph{Grading:} any scalar reward on a
simulator can be satisfied in unintended ways; we found nine such failures in
our own reward and task generator (\S\ref{sec:integrity}).
\emph{Contamination:} many published instruments ship as example files with the
simulator, so a model may reproduce a design it has read rather than reasoned
about; in a pilot audit, 4 of 7 agent episodes retrieved the reference file
through ordinary, permitted tools. \emph{Degeneracy:} a generated family can
look discriminating while admitting one fixed answer, or one rule that copies
numbers from the prompt, that solves most instances, so its pass rates measure
lookup rather than design; we learned this twice (\S\ref{sec:training}).

We introduce \ngym{}, an executable, physically verifiable environment for LLM
agents designing neutron instruments (\S\ref{sec:environment}). Agents act only
through 22 validating tools exposed over the Model Context
Protocol~\citep{anthropic2024mcp}; the environment simulates what they build and
grades it on a level-resolved ladder (L1 syntax, L2 runtime, L3 structure, L4
science) with no LLM judge. Procedural families supply unlimited design
instances with calibrated targets and disjoint held-out parameter regimes; its
curated slice, \mcbench{} (\S\ref{sec:benchmark}), adds reproduction and
improvement tasks built from published instruments, 16 scored per model and
arm, behind memorization probes, perturbed variants and two never-public
held-out instruments. Every scored task fixes the instrument's layout: the
benchmark asks for a published layout rebuilt from its specification, the
families for the numeric parameters of a fixed layout, and no scored tier asks
for a topology choice (the open-design tier is unscored).

The paper follows what happened when we used it. On the benchmark, real
instruments can be reproduced from their specifications, but not reliably:
the best model, Claude Sonnet~5, rebuilt 7 of the 16 tasks in a single
completion and 5 through the tool loop, no episode retrieved a reference, and
no agent met an improvement target (\S\ref{sec:evaluation}); the open-weights
Qwen models \citep{yang2025qwen3} produced no parseable instrument at all
without the tools. Training changed that picture. Reinforcement learning on
the environment's reward took Qwen3-8B from 11.3\% to 76.7\% of held-out
instances on a family whose targets come from a hidden design, past an
untrained 32B that is no stronger than the 8B there, and the same recipe, at
a single seed each, did so on three further gated families
(\S\ref{sec:training}). The analysis then asks what that gain is. Removing
the ladder's partial credit costs 60 points; at the agent's simulation budget
a classical optimizer matches the trained model only when handed the
closed-form physics, which the model recovered from reward alone; and frontier
models still solve nearly every instance (\S\ref{sec:analysis}). Getting a trustworthy version of that result
was the hard part: supervised fine-tuning on the model's own successes
degraded it three times, and four task designs failed the gate because a
no-model baseline solved them with a fixed answer, a rule copying limits from
the prompt, or a whole class of designs (\S\ref{sec:integrity}).

\paragraph{Contributions.}
\begin{itemize}
  \item \textbf{Tasks worth training on.} An admission protocol for procedurally
  generated scientific tasks: three no-model probes (the best fixed answer,
  the best rule that reads the prompt, and every instance's solution applied
  to every other), decided on a one-sided confidence bound, plus a
  concentration diagnostic for the class-level shortcuts the bound cannot see.
  Nine documented reward failures motivated it; four of our own designs failed
  it.
  \item \textbf{\ngym{} and \mcbench{}.} An executable environment for neutron
  instrument design, to our knowledge (search through September 2026) the
  first, with validated construction, compiled execution at about 30 rollouts
  per second per CPU core, a level-resolved reward ladder with a
  reference-free phase-space bound on its maximization families, and
  calibrated procedural families; and its curated slice of published
  instruments, held behind memorization probes, never-public held-out
  instruments, perturbed variants and a sandbox.
  \item \textbf{What training on it shows.} RL takes an 8B model past an
  untrained 32B on four gated families and, from reward alone, to within an
  insignificant four points of a classical search handed the closed-form
  physics; the ladder's partial credit is what makes that trainable, and
  imitation of the model's own successes degrades it.
\end{itemize}
\section{Related Work}
\label{sec:related}

Language-model agents that design physical apparatus against a simulator are
no longer new, and the closest systems run the loop we run.
\citet{kharel2026photonics} let agents propose, simulate and revise photonic
designs against fabrication and physical-consistency gates, and argue the
recipe applies wherever a simulator and acceptance criteria exist; GRACE
\citep{hill2026grace} reproduces and improves Geant4 setups under
physics-motivated utility functions; OptiAgent \citep{geng2026optiagent}
trains on a gated multi-level reward for optical design whose top level is an
LLM judge, where our ladder has none. Detector design has a simulator reward
with RL alone \citep{qasim2024rldesign} and with LLMs proposing designs that
RL then refines \citep{zoccheddu2026llmdesign}, and benchmarks of agents on
scientific simulators and experimental design are numerous
\citep{gandhi2025boxinggym,kumar2026mdgym,shen2026sciagentgym,chi2026frontiereng,
liu2026stargazer,chen2025scienceagentbench}. Nearer our domain, LLM agents
already operate beamlines and analyze scattering data
\citep{chen2026xrayscientist,vandervleuten2025envtrace,li2026rongzai,
xiao2026neudiff,robledo2025sansmcstas,ding2026sasagent}, but none designs
neutron instruments,
whose classical optimizers predate LLMs \citep{bertelsen2017guidebot}. We do
not claim the loop. What neutron optics adds is a conservation law that bounds
intensity with no reference (\S\ref{sec:environment}), Monte Carlo observables
whose noise forces per-observable tolerances, and matching tasks that no
open-weights policy solves without measuring (\S\ref{sec:analysis}).

The training method is likewise standard, and we claim no novelty for it.
Rejection sampling \citep{dong2023raft,zelikman2022star} is the baseline that
fails here; simulator- and verifier-grounded RL
\citep{shi2026mdagent2,cai2026rlvp,zeng2026rlve} is the method that works.
RLVP \citep{cai2026rlvp} is the nearest: it post-trains with the structure
our ladder uses, validity gates and then a continuous physics score, for
PDE-solver code, and reports the result types we report, one policy over
several families, transfer to a held-out family, and a small post-trained
model against a prompted frontier model. On that last comparison our result
is the opposite, the frontier models staying ahead (\S\ref{sec:analysis}).
Calibrating difficulty to the agent follows \citet{zeng2026rlve} and
\citet{guo2025genenv}.

What this paper adds sits around the training rather than inside it. Agents
exploit tool-use rewards and game verifiers
\citep{thaman2026rhb,pan2024icrh,helff2026gaming}, verifiers can be fuzzed
\citep{ray2026fuzzing,roth2026hackverifiable}, agent benchmarks have been
broken outright \citep{wang2026brokebenchmarks}, and procedural generation is
an established contamination defence
\citep{srivastava2026beyondbench,bentham2026infinitesciencegym}. We take
those warnings as a design requirement rather than a caveat: every family
must survive no-model probes before any pass rate is read as capability, and
four of our own designs did not. The contrast with environments whose
feedback a language model produces \citep{envsimbench2026} sharpens what a
physical simulator buys and costs: our reward cannot hallucinate, because it
is a ray trace, but we must ship a real simulator, and the failures we
engineered against were degenerate tasks, not fabricated observations.
Appendix~\ref{app:related} lists further related work.
\section{The NeutronGym Environment}
\label{sec:environment}

\begin{figure}[!ht]
\centering
\includegraphics[width=\textwidth]{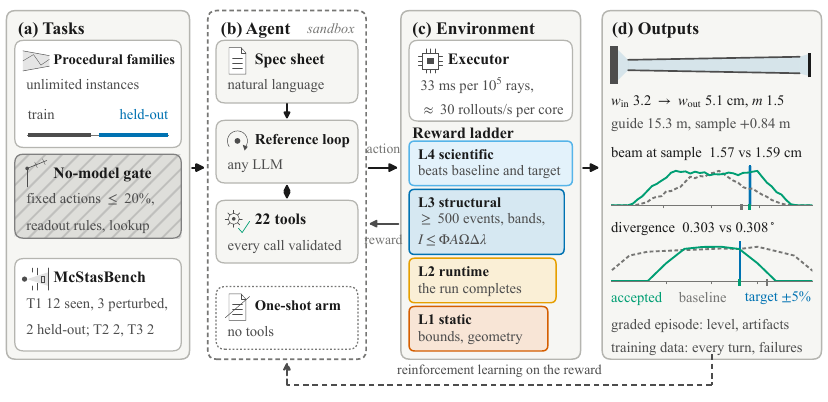}
\caption{\ngym{}. \textbf{(a)} Tasks come from procedural families or from
\mcbench{}, the curated slice; \textbf{(b)} the agent works only through
validated tools in a sandbox, and \textbf{(c)} the environment runs what it
builds and grades it on a level-resolved ladder. \textbf{(d)} One episode: the
instrument the agent arrived at, and the beam profile (horizontal position)
and divergence (horizontal angle) McStas returns for it (solid) against the
baseline (dashed) and a hidden design's targets, shaded at $\pm5\%$; $m$ is
the supermirror coating index.}
\label{fig:setup}
\end{figure}

\subsection{McStas for neutron instrument design}
Everything in \ngym{} rests on one fact about the simulator. An instrument in
McStas~\citep{willendrup2020mcstas} is a text file: a beam-ordered list of
components (source, optics, sample, monitors), each placed in space and
parameterized. McStas compiles the file to C, traces neutron rays through it,
and reports what each monitor would see, with Monte Carlo error bars. That
report is the reward (Figure~\ref{fig:setup}). No model is ever asked whether
a design is good; the environment measures the design the agent actually
built, which is why the score cannot be argued with.

The agent never has to write that file and never sees a shell. It works
through 22 tools over MCP wrapping McStasScript~\citep{mcstasscript}
(Appendix~\ref{app:tools}): it can look up any of the 374 shipped components,
place one, run a simulation, and read back per-monitor summary statistics
rather than raw arrays. Every mutating call is checked against component
metadata as it arrives, so an unknown component or a misspelled parameter
fails at the tool call, with the nearest valid alternative suggested, instead
of surfacing later as a compiler error. Raw \instr{} import and export remain
as an escape hatch.

Trainability comes down to speed. The standard \texttt{mcrun} wrapper spends
about 2.4\,s per run on translation and compilation whatever the ray count,
so each family compiles once, keyed on a hash of its comment-stripped source,
and every instance is a parameter assignment executed directly on the binary:
33\,ms median at $10^5$ rays, about 30 rollouts per second per CPU core.

\subsection{Grading an action: the reward ladder}
An action is graded by how far it gets, not only by whether it works
(Figure~\ref{fig:setup}c). Four levels are checked in order, and the first
three are engineering. \textbf{L1} needs no simulation: the action must
assign exactly the family's free parameters, all finite and inside their
bounds, and pass the family's geometric specification checks. \textbf{L2}
asks that a short run at reduced ray count completes. \textbf{L3} runs the
environment's protocol, with ray count and seed fixed per instance and never
chosen by the agent, and requires the graded monitor to hold at least 500
events; on the two maximization families it also requires the observables to
stay inside band constraints relative to the baseline and the intensity to
respect the phase-space bound below (the matching families carry neither). An
instrument that clears L3 exists, runs, and produces usable statistics.

\textbf{L4} is the physics, and what it asks depends on the family. A
\emph{maximization} family wants the figure of merit (FOM) above the
instance's calibrated target, $r = \mathrm{FOM} / (\mathrm{baseline} \times
\text{target fraction}) > 1$ strictly; grading is deterministic at the fixed
seed, so resubmitting the baseline scores exactly $1$ and is not an
improvement. A \emph{matching} family wants every matched observable within
its tolerance of the instance's target, with $r$ the tolerance over the worst
relative error, so $r > 1$ again means a pass and noisier observables can
carry wider tolerances.

The reward follows the ladder: $0.25$ per level through L3, then
$0.75 + 0.25\min(r, 2)$ once L3 holds, whether or not the bar $r > 1$ is met,
capped at $1.25$. It rises with depth, then with margin, and that shaping is
load-bearing: replacing it with pass/fail costs 60 points
(\S\ref{sec:analysis}). Gate-then-score is the structure verifier-grounded
post-training uses elsewhere \citep{cai2026rlvp,geng2026optiagent}.

\paragraph{Phase-space bound.} One check needs no reference. Passive neutron
optics cannot increase phase-space density \citep{bertelsen2016phasespace},
so the intensity at a monitor is bounded by $I \le \Phi A \Omega
\Delta\lambda$, source brilliance times the monitor's area, solid-angle and
wavelength acceptance, and an action above $1.10\times$ that bound fails L3.
It is derived for the two maximization families only
(Appendix~\ref{app:bound}); the four matching families, on which every
training number is measured, carry none, since their targets are a hidden
design's observables and an inflated intensity wins nothing.

\subsection{Task families and calibrated targets}
A \emph{family} is one instrument layout held fixed while everything numeric
varies: some parameters describe the instance the agent is handed, the rest
are the design knobs it has to set. An \emph{instance} is a deterministic draw
from \emph{(family, split, index)}, and the held-out split draws at least one
context parameter from an interval disjoint from training, so a held-out
instance is a regime the policy never saw, not a fresh seed. Two of our six
families ask the agent to \emph{maximize} a figure of merit. Four ask it to
\emph{match} targets: the observables a hidden design produced at the protocol
seed, the design itself never shown, with calibration discarding any instance
whose targets the baseline already meets. That distinction turns out to
decide whether a family can be shortcut without designing anything
(\S\ref{sec:integrity}); Appendix~\ref{app:families} gives each family's
parameters, specification, held-out regime and status.

A target means nothing except relative to what is achievable on that
instance. We therefore calibrate each with a constraint-filtered search (30
random valid samples, then a local pattern search) that scores every
candidate exactly as an agent is scored, at the protocol seed, so target and
agent share the same Monte Carlo noise; the target is $0.85\times$ the best
figure of merit found. We learned this the hard way: an earlier
version set targets against a deliberately undersized baseline, and untrained
models passed 80--93\% of the guide family's held-out instances. That was a
ceiling, not a task.
\section{McStasBench}
\label{sec:benchmark}

\subsection{Tasks, grading and controls}
Where the procedural families are made up, \mcbench{} asks for real
instruments, in three ways. A \textbf{T1 (reproduce)} task hands
the agent a specification sheet (parameters, component order and types,
monitors to report) and asks it to rebuild the instrument; the sheet is
generated from a machine-verified reference \instr{}, so it is faithful by
construction, and the reference passes its own task at a fresh seed. A
\textbf{T2 (improve)} task hands over a parameterized baseline and a target
$80\%$ of the way to a classical optimum. A \textbf{T3 (open design)} task
asks for an instrument from scratch and is released unscored. Over 12
published instruments of five classes, 16 tasks are scored per model and arm.

What a pass means deserves stating plainly. Grading runs candidate and
reference under one protocol and needs every graded observable (intensity,
beam width, spectral centroid and width at the named monitors) inside
tolerance, so a T1 pass certifies beam delivery, not class-defining
performance such as a diffractometer's resolution. And because the sheet
comes from the reference and McStas is deterministic at the graded seed, a
faithful rebuild reproduces the reference exactly: 32 of the 39 scored
seen-tier passes and all 12 held-out passes match it on every observable
exactly. In practice a T1 pass is an exact reproduction, and the
tolerance band (Appendix~\ref{app:tasks}) rarely decides one.

The harder question is whether a pass is design or recall, and three controls
separate the two. Memorization probes record what each model can write from
memory before it sees the benchmark; Claude Sonnet~5 reproduces one of the 16
scored references, the others fewer or none, and it fails that task in both
arms, so memory accounts for none of its passes. Two held-out instruments have
references we authored from published designs and never released. And a
perturbed variant of a seen task is kept only on a \emph{pair proof}, in which
the perturbed reference passes its own task where the canonical one fails it,
so a memorized file fails exactly where a design that follows the sheet
passes. Underneath all three sits the sandbox: in benchmark mode the tool
server refuses example retrieval and file loads outside the episode, the loop
has no shell or file access, and an automated transcript audit
(Appendix~\ref{app:controls}) invalidates any episode that touched a
reference.

\paragraph{Protocol.} All numbers use the \ngym{} reference loop, a
model-agnostic chat-completions client over the tool server; a one-shot arm
asks for the whole \instr{} file in a single completion, without tools, graded
identically. Episodes are pass@1 at temperature 0 with a 50-turn cap, each API
model pinned to one serving provider
(Appendices~\ref{app:tasks}--\ref{app:prompts}).

\subsection{A small-scale benchmark test}
\label{sec:evaluation}
\begin{table}[!h]
\caption{\mcbench{} results, one episode per task; no per-model difference is
significant at this size. ``Dominant failure'' is where most failed loop
episodes stop on the ladder. Maverick's held-out episodes ran on two serving
routes (Appendix~\ref{app:models}) and all four failed.}
\label{tab:bench}
\centering\small
\begin{tabular}{@{}lcccccl@{}}
\toprule
& \multicolumn{2}{c}{Seen (of 16)} & & \multicolumn{2}{c}{Held-out (of 2)} & \\
\cmidrule(lr){2-3}\cmidrule(lr){5-6}
Model & Loop & 1-shot & & Loop & 1-shot & Dominant failure \\
\midrule
Claude Sonnet 5        & 5 & 7 & & 2 & 1 & L1 does not compile \\
Claude Haiku 4.5       & 4 & 4 & & 2 & 0 & L4 wrong physics \\
Gemini 3.6 Flash       & 4 & 6 & & 2 & 1 & L0 never finished \\
Gemini 3.5 Flash-Lite  & 2 & 1 & & 1 & 0 & L0 never finished \\
Llama 4 Maverick       & 0 & 2 & & 0 & 0 & L0 never finished \\
Qwen3-32B              & 3 & 0 & & 2 & 0 & L0 never finished \\
Qwen3-8B               & 1 & 0 & & 1 & 0 & L0 never finished \\
\bottomrule
\end{tabular}
\end{table}
\mcbench{} is small by construction: 16 scored tasks, one episode per task and
arm, seven models (Table~\ref{tab:bench}; Appendix~\ref{app:matrix}). That is
enough to ask whether a scaffolded model can rebuild a published instrument
from a specification and not enough to rank models, since the Wilson 95\%
interval on 7 of 16 spans 23--67\%, and we read it accordingly. Failed
episodes are placed on the ladder of \S\ref{sec:environment} by what they
produced: \textbf{L0} no finished instrument, \textbf{L1} does not compile,
\textbf{L2} dies during the run, \textbf{L3} a graded monitor missing or
starved, \textbf{L4} gradable but outside tolerance.

Three things came out of it. First, the controls held: no episode in any arm
retrieved a reference through the tool surface. Second, reproduction is
possible but unreliable. The best model passes 7 of 16 tasks one-shot (5 of
the 12 instruments) and 5 through the loop (3 instruments), and the Qwen
models show what the tools are worth, passing nothing one-shot because they
never emit a parseable file. Third, the improvement tier is unsolved: no agent
in either arm met the target of the one scored T2 task or of its development
twin, each set $80\%$ of the way to a classical optimum that improves the
baseline by $2.95\times$ and $1.47\times$, which budget-matched comparisons
of LLMs against classical search have found elsewhere
\citep{rodrigues2026hpo}.

\paragraph{The scaffolds fail in different places.} Loop and one-shot pass
counts do not differ for any model (Fisher exact $p \ge 0.22$), so we make no
scaffold claim, but their failures differ in kind: one-shot failures pile up
at L1, where the file does not compile, and loop failures at L0, where the
agent never finishes an instrument.
Over the 14 model--instrument cells of the held-out tier the loop passed 10
against 2 one-shot ($p = 0.006$); those two instruments are also easier than
the seen tier, where the loop passed 19 of the seven models' 112 cells
($p < 10^{-4}$), a confound two instruments cannot settle.
\section{Training in \ngym{}}
\label{sec:rl}

Training is where the paper's two halves meet: a task that cannot be
shortcut, and a method that learns from its reward. Getting the first took
longer. We describe the failures that shaped the tasks, the training that
followed, and what bounds the result.

\subsection{Making the tasks trainable}
\label{sec:integrity}
\label{sec:setup}
We attacked every scored signal before trusting it and found nine ways to
score without designing anything (Appendix~\ref{app:redteam}). Four changed
the environment's design, and two of them set all that follows: a rule that
read the answer off the prompt, and a family that a whole class of designs
could solve.

\paragraph{The admission gate.} The rule we arrived at is a gate. \ngym{}
accepts a family only after no-model probes of three kinds: fixed actions,
rules that read quantities stated in the prompt (including hand-coded physics
inversions), and, for target-matching families, every instance's own solution
applied to every other instance. A family fails if the baseline passes
anywhere, if a copy-type rule succeeds, or if the best constant's pass share
exceeds 20\% on the one-sided 95\% Clopper--Pearson upper
bound~\citep{clopper1934}, not the observed share; the 20\% level is a choice,
and one admitted family clears it by two points. A physics inversion that
passes is reported beside the trained model as a reference arm, since it needs
the physics the task tests (Appendix~\ref{app:probes}).

\paragraph{What a fixed-answer gate cannot see: concentration.} The gate has a
blind spot: it bounds what one \emph{fixed} design can pass, not what a
\emph{class} can. Our first bender family passed it at 10.7\% while a quarter
of its instances fell to any sufficiently transparent design, because their
hidden design filtered nothing; that is a class, not a point, so no single
candidate ever scored high. We therefore also record the design behind each
passing episode and report how concentrated those designs are, a diagnostic
read by hand rather than a gate. It exposed the broken family (42 distinct
designs in 141 passes, one reused 17 times), while the admitted families read
225 distinct in 230 passes (guide match), 138 in 140 (TOF chopper), 115 in 133
(SANS match) and 63 in 227 (bender, both of whose targets follow from one
cutoff wavelength). A class-level criterion is an open problem we name rather
than solve.

\paragraph{Rejected designs, and the family we train on.} Reinforcement
learning on the original guide family took the 8B from 18.0\% to 98.7\%,
above the untrained 32B at 77.3\%, and looked like the result we wanted. It
was not: the prompt prints the two
binding limits, and a no-model rule copying them passes 96.3\%,
indistinguishable from the trained model (147 and 149 of 150 paired instances,
$p = 0.50$). The fixed-action gate could not see it, because the rule reads
the prompt. Four designs failed the gate, three discarded and one kept
for evaluation only: the undersized-baseline guide version
(\S\ref{sec:environment}), this repaired guide family, a scattering family
whose readout rule passes 56\% (the one kept), and the first bender family. We
arrived at \emph{guide match}: its targets are the spot size and divergence of
a \emph{hidden} per-instance design, matched within $\pm5\%$, so nothing in
the prompt can be copied and no answer transfers. Its best fixed action passes
4.0\% (upper bound 7.7\%); a closed-form optics inversion passes 27.7\% and is
the reference arm.

\paragraph{Training and evaluation protocol.} \emph{Imitation} keeps the
untrained model's successful
episodes and fine-tunes on them with LoRA~\citep{hu2022lora}, in the style of
rejection sampling~\citep{dong2023raft}. \emph{Reinforcement learning} uses
GRPO~\citep{shao2024deepseekmath} on the environment's reward: the training
states are every turn of the untrained model's episodes on 300 training
instances, failures included ($\approx$2{,}810 states); each update samples 8
completions per state, scores them with the ladder, and applies a
group-normalized policy gradient with a KL penalty to the frozen base (LoRA
rank 32, about 3.5 hours on one A100; Appendix~\ref{app:training}). Every
number below is measured on 300 held-out instances of a family (150 for the
out-of-range split) at temperature 0 with a 10-turn budget,
paired with the baseline on the same instances by exact
McNemar~\citep{mcnemar1947} (Appendix~\ref{app:stats}), on a slice that played
no part in choosing tolerances or, except where we say so, checkpoints;
tolerances were fixed by measuring the untrained model before training
(Appendix~\ref{app:families}).

\subsection{Results}
\label{sec:training}
\begin{figure}[!h]
\centering
\includegraphics[width=\textwidth]{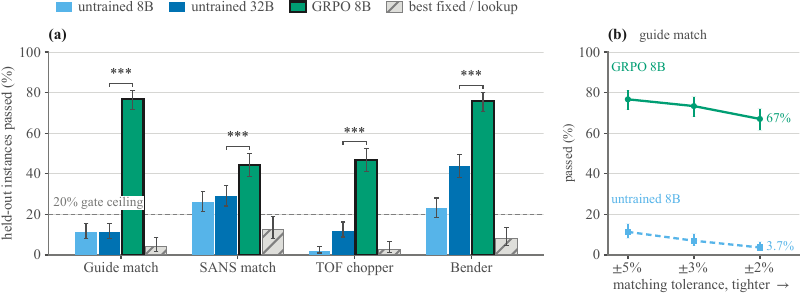}
\caption{Pass rates on held-out instances ($n=300$; $150$ for the no-model
probe) with Wilson 95\% intervals, covering instance sampling for a fixed
policy, not training-seed variation (8 points) or
grading-seed fragility (\S\ref{sec:analysis}). \textbf{(a)} The trained 8B
passes the untrained 32B on all four gated families and no fixed-answer or
lookup baseline comes close; $^{***}$: $p<10^{-3}$, paired exact McNemar against the untrained
32B (discordant 208 vs.\ 12, 89 vs.\ 43, 116 vs.\ 12, 125 vs.\ 29). Guide
match is replicated at a second seed; the other three are single runs.
\textbf{(b)} Guide match under a tightening tolerance: the trained policy's
designs sit inside the tolerance, the untrained one's on its edge.}
\label{fig:families}
\end{figure}

\paragraph{The headline.} On guide match, GRPO takes Qwen3-8B from 11.3\% to
76.7\% of 300 held-out instances (34 to 230 passes; discordant 213 vs.\ 17
against the untrained 8B, 208 vs.\ 12 against the untrained 32B). A second
training seed gives 69.0\%, and the two seeds differ ($p = 0.043$).

\paragraph{Imitating your own successes does not work.} The pre-registered
imitation run made the model \emph{worse}:
40.3\% to 31.3\% on 300 paired held-out instances (discordant 20 vs.\ 47,
$p = 0.0013$), and two further runs on the scattering family regressed as well
(Table~\ref{tab:imitation}). Cloning your own passing turns teaches a move
that was right in the states you already handled and is a trap everywhere
else; on-policy RL trains on decisions from every state, failures included.

\paragraph{The recipe holds on three further families, at one seed each.}
We built three more matching families and gated each before training: \emph{SANS match}, the
collimation of a small-angle neutron scattering instrument, matching beam
widths at the sample and the beamstop; \emph{TOF chopper}, which leaves
geometry altogether, a time-of-flight chopper pair's frequency, phase and
opening angle set to deliver a required mean wavelength and spread; and
\emph{bender}, a curved guide that transmits long wavelengths and suppresses
short ones, with targets on the transmitted band's centre and width.

Paired on one slice against the untrained 32B, the trained 8B wins every
family at $p \le 8\times10^{-5}$ (Figure~\ref{fig:families}a). We do not
average the four: gains against the untrained 8B run from 18 to 65 points,
three of the four families rest on a single training run, and SANS match is
clean by a slim margin, its best fixed design passing 12.7\% (95\% bound
18.0\%) against the 20\% ceiling. On that family, the only one where two
checkpoints were scored, the kept checkpoint read 51.0\% on the slice used to
choose it and 44.3\% on a slice chosen for nothing, the number
Figure~\ref{fig:families} reports.

\paragraph{One policy can cover all four families, but the recipe is unstable.}
A single LoRA trained from the base model on the pooled decision points of all
four families, at one specialist's budget of 220 updates split four ways,
reaches 43--64\% on every held-out family (Table~\ref{tab:joint}). The
per-family pattern should not be read, however. Repeating the three-family
configuration with nothing changed but the seed produced \textbf{2.3\% on TOF
chopper, where the untrained rate is 2.0\%, against 55.0\% for the first run},
and training telemetry alone shows why: about a quarter of sampled groups
carried any gradient signal, against roughly half in the run that worked
(Appendix~\ref{app:joint}). That is a negative result about multi-task RL,
and since its cause lies in the optimization rather than the pooling, it also
bounds our confidence in the single-seed specialists.

\subsection{Analysis}
\label{sec:analysis}
\begin{table}[!h]
\caption{Attempts to explain the result away. On guide match the RL-trained 8B
passes 76.7\% of 300 held-out instances; each row is a candidate account of
that number, read against the reference in its own row ($\Delta$ in points).
All search rows are coordinate descent, the best of the three optimizers we
tried (Nelder--Mead, random search, coordinate descent) at both budgets, and
differ only in where the search starts; the seeded search's 4-point lead does
not separate at this size (\S\ref{sec:analysis}).}
\label{tab:buys}
\centering\small\setlength{\tabcolsep}{3pt}
\begin{tabular}{@{}llccr@{}}
\toprule
& Family / slice & Pass rate & Read against & $\Delta$ \\
\midrule
\multicolumn{5}{@{}l}{\emph{Something simpler did it}} \\
\quad Search from the baseline, agent's budget & guide match & 13.3\% & 76.7\% RL & $-63$ \\
\quad Search from a random start, agent's budget & guide match & 14.3\% & 76.7\% RL & $-62$ \\
\quad Search from the baseline, $3\times$ the budget & guide match & 73.7\% & 76.7\% RL & $-3$ \\
\quad Hand-coded physics rule, no search & guide match & 27.7\% & 76.7\% RL & $-49$ \\
\quad Physics rule, then search, agent's budget & guide match & 81.0\% & 76.7\% RL & $+4$ \\
\quad Frontier models, zero-shot & guide match, first 100 & 98--99\% & 76.7\% RL & $+22$ \\
\addlinespace
\multicolumn{5}{@{}l}{\emph{It is an artifact of the evaluation conditions}} \\
\quad Beyond the training parameter ranges & guide match, $n=150$ & 45.3\% & 17.3\% untrained & $+28$ \\
\quad At a $\pm2\%$ tolerance & guide match & 67.0\% & \phantom{0}3.7\% untrained & $+63$ \\
\quad A family it never trained on & guide (rejected) & 34.0\% & 18.0\% untrained & $+16$ \\
\addlinespace
\multicolumn{5}{@{}l}{\emph{What does account for it}} \\
\quad Reward ladder removed, pass/fail only & guide match & 16.7\% & 76.7\% with the ladder & $-60$ \\
\bottomrule
\end{tabular}
\end{table}
Table~\ref{tab:buys} collects every attempt we made to explain the result
away: something simpler did it, the conditions flatter it, or the ladder did
it. This section reads them in turn.

\paragraph{On guide match, reward alone reaches what the physics supplies by
hand.} Over one held-out slice the arms run from the untrained open-weights
models through classical search from the baseline and a hand-coded optics
inversion to the trained model and a search seeded by that inversion, with
frontier models at 98--99\% on the slice's first 100 instances, where the
trained model scores 76 (Appendix~\ref{app:guidematch}). Training recovers
about three quarters of the gap to the frontier and does not close it; the
frontier models are zero-shot where the trained model saw 300 training
instances. Coordinate descent started from the closed-form inversion reaches
81.0\% at the agent's budget against the trained model's 76.7\%, a gap that
does not separate at this size (discordant 59 vs.\ 46, $p = 0.24$), while
from a random valid start it reaches 14.3\%, no better than from the baseline
($p = 0.80$): the physics, not the start, carries the information, and the
model recovered it from reward alone. The optimizers act on the numeric parameters directly and
the seeded one is handed the physics; the agent reads the parametrization
from a task sheet, must relate it to the observables itself, and reaches the
simulator only through tool calls. Nor is the gain recall: every untrained
episode reaches L3, so the gain is all at L4; on guide match and TOF chopper
no 8B or 32B policy, trained or untrained, passes on the first turn (SANS
match: 2 of 133; bender: 18 of 227); and almost every pass is distinct (225
of 230). The two seeds' pass sets overlap (agreement 0.68, $p = 0.025$).

\paragraph{Stress tests, and one fragility.} Tightening the tolerance from $\pm5\%$ to $\pm2\%$ widens the
trained-to-untrained ratio from $6.8\times$ to $18\times$, and the ordering
survives the out-of-range split of Table~\ref{tab:buys}
($p = 7\times10^{-7}$). One fragility counts against the headline: re-grading
all 230 passing designs at three fresh seeds, 80\% hold at every seed and
88\% of re-runs hold, while 5\% fail at all three; the misses cluster at the
boundary, tolerance-edge fragility rather than wrong designs.

\paragraph{The reward ladder is load-bearing.} Replacing the level-resolved
reward with pass/fail only, all else fixed, collapses the result to 16.7\%
(discordant 196 vs.\ 16, $p = 1.5\times10^{-40}$), and the mechanism is
visible during training: with sparse reward a median of 0.33 of the 8 sampled
groups per update contained any reward spread, against 5--7 with shaping, so
nearly every group contributed no gradient. Dense partial credit is what makes
this task reliably trainable at 8B scale in our runs, with three bounds: the
ablation is one seed against one; the same gradient-starved signature appeared with the
ladder present in the collapsed joint run (\S\ref{sec:training}), so the
ladder is necessary, not sufficient; and pass/fail removes level and margin
credit together, so the ablation cannot apportion the 60 points.

\paragraph{The gain partly transfers, and more training is not better.} On the
rejected guide family (\S\ref{sec:integrity}), which it never trained on, the
guide-match model nearly doubles the untrained rate (discordant 76 vs.\ 28,
$p = 3\times10^{-6}$) and its rejections before simulation fall from 73 to 20;
since a prompt-copying rule passes 96.3\% of that family, this shows the habit
RL teaches (propose a valid geometry, then refine), not design skill. Training
longer hurts: checkpoints at 120, 220 and 340 updates score 50.0\%, 73.0\% and
56.7\% on another held-out slice as training reward keeps rising.

\paragraph{What was pre-registered, and one measurement defect.}
Pre-registered: the first supervised run and the matched-compute, frontier and
ablation experiments; the rest is exploratory. Before 2026-09-16 the
environment reported the per-turn figure of merit to the agent as a fraction
of the target but labelled it as relative to the baseline; that defect covers
the supervised runs (Table~\ref{tab:imitation}) and the guide GRPO run of
\S\ref{sec:integrity}, none re-measured, and no matching-family or benchmark
number.
\section{Limitations and Conclusion}
\label{sec:discussion}

\paragraph{Benchmark scale.} The curated slice is small: 16 tasks at one
episode each cannot rank models or separate scaffolds; it tests whether
reproduction is possible at all. Its two held-out
instruments are simpler than the seen median and not facility-validated, so
they separate neither novelty from difficulty nor model failure from
reference error.

\paragraph{Task coverage.} No scored task asks for a topology choice: the
benchmark grades beam delivery at the monitors a specification names, and the
families set two or three parameters of a fixed layout, where every training
number lives. The procedural side is narrow in kind, its four matching
families were built after we found the shortcuts, the simulation is idealized
(ideal monitors, no background, gravity off), and the probe suite is
incomplete by construction: the gate bounds what one fixed answer can pass,
not a class, and admitted a quarter-degenerate bender family
(\S\ref{sec:integrity}).

\paragraph{Strength of the training evidence.} The training results are
exploratory: one seed per configuration except two, and a multi-family
configuration bimodal across seeds (\S\ref{sec:training}). The trained model is an 8B on two- and
three-parameter tasks, and \S\ref{sec:analysis} bounds what it has: below
frontier models on the same slice, matched at equal budget by a search handed
the physics, weaker outside its training ranges, seed-fragile in a fifth of
passes; a second seed per family is the next step.

\paragraph{Conclusion.}
\ngym{} makes neutron instrument design, at the level of a fixed layout's
parameters, executable and gradeable by physics: seven models built
instruments in it without retrieving a reference, and RL on its reward lifted
an 8B past an untrained 32B on four gated families, three at a single seed.
Only no-model baselines separate a task that measures design from one that
rewards a shortcut. We release it all.

\clearpage
\subsection*{AI Use Statement}
We used an AI coding assistant (Claude, via Claude Code) extensively. It helped
implement the tool server, environment, and evaluation harness; run and monitor
experiments; analyze results; and draft and revise this manuscript. All code is
covered by an automated test suite. Every number in the paper is regenerated by
scripts from committed per-episode evidence rather than transcribed by hand, and
the authors checked the reported numbers against that evidence. The
experimental design, pre-registration decisions and scientific claims are the
authors' responsibility. Models from the same provider as the assistant are
among those evaluated; all models were evaluated through the same scaffold and
grader.

\subsection*{Ethics Statement}
This work involves no human subjects or personal data. Neutron instrument
design has no direct dual-use concern beyond that of the published simulator.
All models were accessed under their providers' terms.

\subsection*{Reproducibility Statement}
We release the environment as an installable package with a one-command
evaluation of the procedural slice, the benchmark tasks and reference
instruments, the pinned run configuration (model identifiers, provider pins,
protocol constants), the per-episode evidence for every reported result
(grading report, full transcript and the instrument the agent built), the
training and evaluation records behind every number in \S\ref{sec:training}
and \S\ref{sec:analysis} with the step-level telemetry that diagnoses the
unstable run, and the trained LoRA adapters with checksums. Task
families, calibration and the constant-policy gate are deterministic given
their seeds; the residual nondeterminism is in serving: temperature-0 decoding
under batched inference is not bit-reproducible, and one API model behaved
differently on two providers (Appendix~\ref{app:models}). Three interim claims retracted during the work are
listed in the release with the evidence that retracted them: a
scaffold-superiority claim that did not survive its statistics, an apparent
training gain that a no-model baseline matched, and a task family whose
targets a quarter of its instances did not constrain. Code, tasks and evidence:
\url{https://github.com/ljding94/NeutronGym}.



\clearpage
\appendix
\section*{Appendix}
The appendix is organised in the order the main text needs it: what the agent
can do (\ref{app:models}--\ref{app:tools}), what it is asked to do
(\ref{app:tasks}--\ref{app:prompts}), related work in full (\ref{app:related}),
how we established that the tasks are worth training on and what the
phase-space bound is (\ref{app:redteam}--\ref{app:bound}), how the training and the statistics
were run (\ref{app:training}--\ref{app:stats}), and the full result tables
(\ref{app:matrix}--\ref{app:guidematch}).

\section{Models and Serving}
\label{app:models}

\begin{table}[h]
\caption{Models and serving routes. The eighth row ran one seen-tier episode
(\texttt{HZB\_FLEX}, passed) before being cut on cost and is excluded from
every reported number.}
\label{tab:models}
\centering\small
\begin{tabular}{@{}llll@{}}
\toprule
Model & Access & Serving & Notes \\
\midrule
Claude Sonnet 5 & OpenRouter & provider-pinned (Google) & \\
Claude Haiku 4.5 & OpenRouter & provider-pinned (Google) & \\
Gemini 3.6 Flash & OpenRouter & provider-pinned (Google) & \\
Gemini 3.5 Flash-Lite & OpenRouter & provider-pinned (Google) & \\
Llama 4 Maverick & OpenRouter & DigitalOcean$^\ddagger$ & see \S\ref{sec:evaluation} \\
Qwen3-32B & local vLLM & 4 GPUs, YaRN $\times$4, 98{,}304 ctx & thinking off \\
Qwen3-8B & local vLLM & 1 GPU, YaRN $\times$4, 98{,}304 ctx & thinking off \\
GPT-5.2 Pro & OpenRouter & OpenAI & one episode (pass), cut on cost \\
\bottomrule
\end{tabular}

\smallskip
$^\ddagger$Google until that route was withdrawn; its held-out loop episodes
ran on Google and its held-out one-shot pair on DigitalOcean.
\end{table}
Provider pins are those recorded in the pinned run configuration. The Qwen
models were served with thinking disabled in the chat template, the pinned
non-thinking baseline configuration; API models were called with no reasoning
parameter set, so at provider defaults. Thinking-enabled Qwen arms were not
run, which is a limit on the ``past an untrained 32B'' comparison.

\section{The Tool Surface}
\label{app:tools}
The 22 tools the agent acts through, wrapping McStasScript~\citep{mcstasscript} over MCP. Every
mutating call is validated against component metadata at call time: an unknown
component or parameter is refused with the nearest valid alternative rather than
surfacing later as a compiler error. In benchmark mode
\texttt{list\_examples}, \texttt{get\_example} and \texttt{load\_instr\_file}
are disabled, which is what closes the retrieval route the pilot audit found.

\begin{table}[h]
\caption{The tool surface. Agents never write files or run a shell; these calls
are the whole action space.}
\label{tab:tools}
\centering\footnotesize\setlength{\tabcolsep}{5pt}
\begin{tabular}{@{}lp{9.2cm}@{}}
\toprule
Tool & Purpose \\
\midrule
\multicolumn{2}{@{}l}{\emph{Introspection: the 374 shipped components}} \\
\texttt{list\_components}   & Browse components by category or search term \\
\texttt{describe\_component}& Full parameter table for one component: type, unit, default, required, documentation \\
\texttt{list\_examples}     & Browse the shipped example instruments \\
\texttt{get\_example}       & Full source of one shipped example \\
\addlinespace
\multicolumn{2}{@{}l}{\emph{Construction: validated at call time}} \\
\texttt{create\_instrument} & Start a new empty instrument \\
\texttt{add\_parameter}     & Add an instrument-level parameter settable per run \\
\texttt{add\_declare}       & Declare an instrument-scope C variable \\
\texttt{add\_component}     & Append a component in beam order, with position and rotation \\
\texttt{set\_parameters}    & Overwrite parameters on a placed component \\
\texttt{remove\_component}  & Remove a component; refused if others are positioned relative to it \\
\texttt{get\_instrument}    & Current state: parameters and components in beam order \\
\texttt{list\_instruments}  & Instruments in the registry \\
\texttt{validate\_instrument} & Static check without a run: required-parameter audit and structural rules \\
\addlinespace
\multicolumn{2}{@{}l}{\emph{Execution: asynchronous jobs}} \\
\texttt{run\_simulation}    & Compile and run at a chosen ray count and seed \\
\texttt{job\_status}        & State of a running job, with elapsed time and log tail \\
\texttt{cancel\_job}        & Kill a running simulation \\
\texttt{get\_results}       & Summary statistics per monitor (intensity, error, counts, peak, centre of mass, FWHM), never raw arrays \\
\texttt{get\_monitor\_data} & One monitor in detail, as a rendered plot or a data slice \\
\addlinespace
\multicolumn{2}{@{}l}{\emph{Search}} \\
\texttt{scan\_parameter}    & Linear scan of one parameter \\
\texttt{optimize}           & Classical optimization (scipy, \citealp{virtanen2020scipy}) over declared free parameters \\
\addlinespace
\multicolumn{2}{@{}l}{\emph{Escape hatch}} \\
\texttt{load\_instr\_file}  & Import an existing \instr{} file into the registry \\
\texttt{export\_instr\_file}& Write the generated \instr{} to disk \\
\bottomrule
\end{tabular}
\end{table}

\section{Tasks}
\label{app:tasks}
Table~\ref{tab:tasks} is the full catalogue. Every episode runs at $10^6$ rays
under the task's fixed seed; ``mon/obs'' counts the monitor roles compared
against the reference and the observables across them. Seen T1 specification
sheets are auto-generated from the reference, T2 targets close $80\%$ of the gap
from the baseline to the classical optimum, and T3 is released unscored pending
an expert rubric. The 16 scored tasks are the seen T1, perturbed and T2 rows
less \texttt{guide\_divergence}, the task the harness was debugged on. Three
of the twelve seen references are design studies or idealizations rather than
as-built instruments (\texttt{ESS\_IN5\_reprate} is a repetition-rate
multiplication study, \texttt{HZB\_FLEX} models the instrument after its
upgrade, \texttt{SESANS\_Delft} is an idealised instrument). The one scored
T2 task is a fixed instance of the SANS collimation family, whose generated
instances a readout rule passes 56\% of the time (Table~\ref{tab:probes});
no agent passed the fixed instance, so that rule inflates no reported
number.

\paragraph{The tolerance rule.} Monitors are matched between candidate and
reference by role and position rather than by name, and an observable counts as
matched within
$\max(\text{rtol}\cdot|x_{\text{ref}}|,\ n_\sigma(\sigma_{\text{ref}} +
\sigma_{\text{cand}}))$, so a noisy observable is judged against its own error
bars rather than a fixed fraction. Five conditions zero an episode's score
outright: no instrument produced, a failed translation, a failed compilation or
run, a graded monitor role absent, or fewer than 1000 events in a graded
monitor. Otherwise the score is the fraction of observables inside tolerance,
and a pass requires every one of them.

\begin{table}[h]
\caption{Every \mcbench{} task. Seen T1 references are McStas shipped examples;
held-out references we authored from published designs
\citep{wang2025boya,bilheux2015venus} and never released.}
\label{tab:tasks}
\centering\footnotesize\setlength{\tabcolsep}{4pt}
\begin{tabular}{@{}llllp{3.3cm}@{}}
\toprule
Task & Class & Site & Mon/obs & Note \\
\midrule
\multicolumn{5}{@{}l}{\emph{Seen T1: 12 scored}} \\
\texttt{ESS\_IN5\_reprate} & ToF spectrometer & ESS & 3 / 8 & \\
\texttt{HZB\_NEAT}         & ToF spectrometer & HZB & 3 / 8 & \\
\texttt{ILL\_IN6}          & ToF spectrometer & ILL & 3 / 8 & \\
\texttt{HZB\_FLEX}         & Triple-axis      & HZB & 2 / 6 & \\
\texttt{ILL\_H10\_IN8}     & Triple-axis      & ILL & 2 / 6 & \\
\texttt{ILL\_D2B}          & Powder diffr.    & ILL & 2 / 6 & \\
\texttt{PSI\_DMC}          & Powder diffr.    & PSI & 2 / 6 & \\
\texttt{RTP\_DIF}          & Powder diffr.    & RTP & 2 / 6 & Reactor TRIGA PUSPATI \\
\texttt{ILL\_H15\_D11}     & SANS             & ILL & 1 / 3 & \\
\texttt{ILL\_H512\_D22}    & SANS             & ILL & 1 / 3 & \\
\texttt{ISIS\_SANS2d}      & SANS             & ISIS & 2 / 6 & \\
\texttt{SESANS\_Delft}     & Spin-echo SANS   & Delft & 1 / 3 & \\
\addlinespace
\multicolumn{5}{@{}l}{\emph{Perturbed: 3 scored, each kept only on a pair proof}} \\
\texttt{PSI\_DMC\_pert}     & Powder diffr. & PSI  & 2 / 6 & source distance $\times1.15$; canonical fails on 2d and wavelength intensity \\
\texttt{ISIS\_SANS2d\_pert} & SANS          & ISIS & 2 / 6 & source distance $\times1.15$; same two observables \\
\texttt{HZB\_FLEX\_pert}    & Triple-axis   & HZB  & 2 / 6 & guide curvature $\times1.4$; canonical fails on energy intensity \\
\addlinespace
\multicolumn{5}{@{}l}{\emph{Held-out: 2, shown to each model once}} \\
\texttt{BOYA\_CARR} & Multiplexing spec. & CARR & 2 / 6 & single-channel simplification; not facility-validated \\
\texttt{VENUS\_SNS} & ToF imaging & SNS BL-10 & 2 / 6 & not facility-validated \\
\addlinespace
\multicolumn{5}{@{}l}{\emph{T2 improve: 1 scored}} \\
\texttt{sans\_collimation} & SANS optics  & --- & FOM & two pinhole radii; detector intensity under a beam-width constraint \\
\texttt{guide\_divergence} & Guide optics & --- & FOM & three guide parameters; divergence-monitor intensity. Development task \\
\addlinespace
\multicolumn{5}{@{}l}{\emph{T3 open design: 2, unscored}} \\
\texttt{compact\_sans}    & SANS         & --- & floors & $\le 12$\,m, wavelength CoM 4.5--5.5\,\AA{}, brilliance transfer $\le 1$ \\
\texttt{ballistic\_guide} & Guide design & --- & floors & coating budget \\
\bottomrule
\end{tabular}
\end{table}

\paragraph{The grading contract.} Every T1 task runs at $10^6$ rays under seed
1234 and requires at least 1000 events in each graded monitor. The tolerances
are fixed per monitor role rather than per task, so no instrument is graded on a
bar tuned to it:

\begin{center}\small
\begin{tabular}{@{}llll@{}}
\toprule
Role & Intensity & Shape & Position \\
\midrule
\texttt{2d} (position-sensitive) & rtol 0.20, $3\sigma$ & beam width $x$, $y$: rtol 0.30 & --- \\
\texttt{wavelength}              & rtol 0.25, $3\sigma$ & FWHM: rtol 0.50 & centre of mass: rtol 0.05 \\
\texttt{energy}                  & rtol 0.25, $3\sigma$ & FWHM: rtol 0.50 & centre of mass: rtol 0.05 \\
\texttt{tof}                     & rtol 0.25, $3\sigma$ & --- & centre of mass: rtol 0.05 \\
\bottomrule
\end{tabular}
\end{center}

Intensity is the loosest in relative terms and the only observable with a
statistical term, because it varies most between correct implementations of the
same specification; centre of mass, which applies to the spectral roles, is the
tightest at 5\%, because an instrument that delivers the wrong mean wavelength,
energy or arrival time is wrong whatever its throughput. The position-sensitive
role is graded on intensity and beam widths only, and no class-defining
performance figure (a diffractometer's resolution, a SANS instrument's $Q$
range) is graded: a pass certifies beam delivery. In practice the band rarely
decides one. Sheets are generated from the reference and McStas is
deterministic at the graded seed, so a faithful rebuild reproduces the
reference's monitors exactly: 32 of the 39 scored seen-tier passes over the
seven models and all 12 held-out passes have every graded observable at zero
difference from the reference, and the seven passes with any nonzero difference
lie on two instruments (\texttt{HZB\_FLEX} and its perturbed twin,
\texttt{ESS\_IN5\_reprate}). A task's \emph{mon/obs} count in
Table~\ref{tab:tasks} follows from which roles its specification sheet asks
for.

Five further tasks are never scored: \texttt{templateSANS} and
\texttt{templateTOF}, used for harness bring-up; two SANS pilots
(\texttt{P1}, and \texttt{P3} with a deliberately underspecified sheet); and
\texttt{P2}, the memorization control that asks for \texttt{PSI\_DMC} from
memory without tools.

\section{Contamination Controls and Protocol}
\label{app:controls}

\paragraph{Memorization probes.} Each model is asked, without tools and at
temperature 0, to write each reference from memory; answers are scored by
normalized text similarity and by recall of the component sequence. Claude
Sonnet~5 reproduced one of the 16 scored references
(\texttt{ILL\_H10\_IN8}) and one development-split task, Gemini~3.6 Flash
one, Haiku~4.5 and Flash-Lite none. Sonnet~5 fails \texttt{ILL\_H10\_IN8}
in both arms, so its memorized reference is not among its passes. One outcome
changed between runs despite temperature 0, so probes are recorded per model and
per date rather than treated as a fixed property of a model.

\paragraph{Perturbed variants.} The three kept perturbations are two source
distances scaled by $1.15\times$ and one guide curvature by $1.4\times$
(Table~\ref{tab:tasks} names the observables on which the canonical reference
then fails). The other nine seen tasks admitted no perturbation that was both
detectable (the canonical reference must fail the perturbed task) and
valid, the perturbed reference passing its own.

\paragraph{Held-out instruments.} BOYA is a single-channel simplification of the
multiplexing cold-neutron spectrometer at CARR \citep{wang2025boya}, which
removes the multiplexing that distinguishes it; VENUS is the SNS
time-of-flight imaging beamline \citep{bilheux2015venus}. Both references we
authored from the published designs, no public model of either exists, and
neither is facility-validated. Their specification sheets are not released.

\paragraph{Leak audit.} The transcript audit is automated
(\texttt{leak\_audit.py} in the release), the third layer over the disabled
tools and the shell-free loop. It scans every tool call in an episode for the
example tools and for any file load whose target lies outside the episode's
own tree, records refused attempts (the sandbox held) without invalidating the
episode, and treats a flagged call whose result is missing from the transcript
as a leak, since the sandbox cannot then be shown to have held. A leaked
episode is invalid regardless of score. The audit detects retrieval within the
episode; memorization is the probes' job above.

\paragraph{Protocol.} Episodes run pass@1 at temperature 0 with a 50-turn cap
(the procedural-family protocol of \S\ref{sec:setup} uses 10 turns) and
environment-controlled seeds. Open-weights models run on vLLM
\citep{kwon2023vllm} with YaRN scaling \citep{peng2023yarn} at a 98{,}304-token
context and thinking disabled; Appendix~\ref{app:models} gives the per-model
serving route. Episodes failing before reaching the model are excluded and
re-run, which never re-exposes a held-out task because the model never saw it;
no episode in the reported matrix needed this (0 infrastructure exclusions
over 225 valid episodes). The API spend was \$122.16.

\section{Task Prompts}
\label{app:prompts}
A T1 specification sheet states the instruction, the instrument parameters with
their defaults, the beam-ordered component list with each component's McStas
type, position and parameters, the monitor roles the instrument must contain,
and the parameter values to run at. No target value appears: the sheet is a
reformatted rendering of the reference, so T1 measures faithful reconstruction
through the tool surface, not recovery of a hidden design. Sheets run 42 to 345 lines
across the twelve seen instruments (median 120). Below is the shortest of them,
\texttt{ILL\_H10\_IN8}, with its parameter list and closing run line elided
where marked; the four T2 and T3 prompts follow complete, and then the
procedural task sheet for the guide-match instance of Figure~\ref{fig:setup}.
All are wrapped to the page width, with em dashes and $\pm$ in ASCII, and are
otherwise verbatim.

\paragraph{T1 (reproduce), \texttt{ILL\_H10\_IN8}.}
{\footnotesize\begin{verbatim}
Rebuild the following TAS in McStas from its specification, run it, and
    report the integrated intensity (with error and event count) on
    every monitor.

Instrument parameters (defaults in parentheses):
- KF (5)
- KI (0)
- QM (0.5)
- EN (0)
- verbose (1)
- WM (0.233)
[... 22 further parameters ...]

Beamline, in order (positions in meters; AT [x,y,z] RELATIVE to the
    named component; parameters as given):
- Origin: Progress_bar AT [0, 0, 0] RELATIVE ABSOLUTE ROTATED [0, 0, 0]
-     RELATIVE ABSOLUTE
- Source: Source_gen AT [0, 0, 0] RELATIVE ABSOLUTE ROTATED [0, 0, 0]
-     RELATIVE ABSOLUTE
    parameters: radius=0.10, dist=machine_hkl.l1,
        focus_xw=fabs(WM*sin(machine_real.a1*DEG2RAD)), focus_yh=HM,
        E0=machine_hkl.ei, dE=machine_hkl.ei*0.03, I1=0.5874e+13,
        T1=683.7, T2=257.7, I2=2.5099e+13, T3=16.7, I3=1.0343e+12
- SC1: Collimator_linear AT [0, 0, machine_hkl.l1/4] RELATIVE ABSOLUTE
-     ROTATED [0, 0, 0] RELATIVE ABSOLUTE  (WHEN (ALF1 && BET1))
    parameters: xmin=-WM/2, xmax=WM/2, ymin=-HM/2, ymax=HM/2,
        length=machine_hkl.l1/2, divergence=ALF1, divergenceV=BET1
- Guide_out: Arm AT [0, 0, machine_hkl.l1-0.2] RELATIVE ABSOLUTE ROTATED
-     [0, 0, 0] RELATIVE ABSOLUTE

Your instrument must include at least: a 2D position-sensitive monitor;
    a wavelength monitor.
Run with: KF=5, KI=0, QM=1, EN=0, verbose=1, WM=0.233, HM=0.197, RMH=-1,
    RMV=-1, DM=1.807, NHM=15, NVM=15, WA= [... remaining defaults ...]
\end{verbatim}}

\paragraph{T2 (improve), \texttt{guide\_divergence}.}
{\footnotesize\begin{verbatim}
Improve this neutron guide design.

Baseline instrument (provided as t2_guide_divergence): a 10 m straight
    supermirror guide (entry = exit = 0.012 m square, m = 1.5) starts
    1.5 m from a 10x10 cm source emitting 5.0 +- 0.5 Angstrom; the
    target is a 2x2 cm sample area accepting +-0.5 degrees, 5 cm behind
    the guide exit (Divergence_monitor 'divmon').

Goal: maximize the neutron flux into that acceptance (integrated
    intensity on 'divmon').
You may change ONLY these instrument parameters, within bounds: w_in in
    [0.01, 0.09] m, w_out in [0.01, 0.09] m, m_coat in [1.0, 3.0].
Target to beat: divmon intensity >= 0.001033 n/s at the evaluation
    protocol (ncount 1e+06, fixed seed -- your result is re-verified
    independently).
Report your chosen parameters and the achieved flux with error and event
    count.
\end{verbatim}}

\paragraph{T2 (improve), \texttt{sans\_collimation}.}
{\footnotesize\begin{verbatim}
Improve the collimation of this SANS instrument.

Baseline (provided as t2_sans_collimation): a pinhole SANS -- 6 m
    collimation with two circular pinholes (r_pin1 = 0.005 m at 3 m,
    r_pin2 = 0.005 m at 6 m), monodisperse-sphere sample, beamstop, and
    a 60x60 cm 2D detector ('detector') 3 m behind the sample; lambda =
    6.0 +- 0.05 Angstrom.

Goal: maximize the scattered intensity on the detector.
You may change ONLY: r_pin1 in [0.002, 0.02] m, r_pin2 in [0.002, 0.02]
    m.
Hard constraint (pattern-integrity band): the detector pattern RMS width
    dX must stay within [4.539, 5.548] cm (+-10% of the baseline
    pattern). This catches both resolution destruction (over-wide
    pinholes smearing the rings) AND direct-beam leakage around the
    beamstop (which floods the center and shrinks dX) -- the beamstop
    must keep shadowing the direct beam.
Target to beat: detector intensity >= 0.06171 n/s at the evaluation
    protocol (ncount 1e+06, fixed seed).
Report chosen radii and achieved intensity with error, events, and the
    measured dX.
\end{verbatim}}

\paragraph{T3 (open design), \texttt{compact\_sans}.}
{\footnotesize\begin{verbatim}
Design a compact pinhole SANS instrument for a compact
    accelerator-driven neutron source, from scratch.

Requirements:
- Total instrument length (source to detector) <= 12 m.
- Wavelength 5 Angstrom with a velocity-selector-like band of
-     Delta-lambda/lambda approximately 10% FWHM (model the band at the
-     source).
- Must measure scattering from 100 Angstrom monodisperse spheres (use
-     the standard hard-sphere SANS sample model, 1x1 cm, volume
-     fraction 0.1%, contrast 0.6 fm/A^2) with usable Q coverage
-     reaching Qmin <= 0.006 1/Angstrom.
- Include: source, collimation of your choosing, sample, beamstop, a 2D
-     position-sensitive detector sized to your geometry, and a
-     wavelength monitor.
- Assume a source brightness of 1e8 n/s/cm^2/st/Angstrom (state any
-     further assumptions).

Deliver: the instrument (leave it in the registry), your design
    rationale (collimation choice, distances, detector size vs Q range),
    and the flux on sample and detector with errors and event counts at
    ncount 1e6 with a fixed seed.
\end{verbatim}}

\paragraph{T3 (open design), \texttt{ballistic\_guide}.}
{\footnotesize\begin{verbatim}
Design the neutron transport from a cold source to a distant sample
    position, from scratch.

Requirements:
- Source: 12x12 cm cold source, lambda = 4-6 Angstrom (model with
-     Source_simple or a Maxwellian source; state your brightness
-     assumption).
- Sample position: 30 m from the source; sample area 3x3 cm; the
-     delivered beam must have divergence within +-0.6 degrees (both
-     axes).
- Coating budget: use m > 3 on at most 10 m of total guide length (m <=
-     3 elsewhere) -- justify where you spend it.
- Topology is YOUR choice: straight guide, tapered, ballistic
-     (expanding-straight-focusing), or elliptic; single guide or
-     segmented.
- Include diagnostic monitors after each stage and a divergence monitor
-     plus PSD at the sample position.

Deliver: the instrument (leave it in the registry), the topology
    rationale, and the figure of merit -- flux into the 3x3 cm, +-0.6 deg
    acceptance at the sample -- with error and event count at ncount 1e6,
    fixed seed. Compare at least two topologies and report both numbers.
\end{verbatim}}

\paragraph{Procedural family, \texttt{guide\_match-heldout-000300}.} Every
family sheet has this shape: the fixed context, the free parameters with their
bounds and the baseline point, and the targets, which are filled in at reset
from the hidden design. The agent is told the parametrization; what it must
supply is the mapping from parameters to observables, by measuring.
{\footnotesize\begin{verbatim}
Instrument design task guide_match-heldout-000300.

Shape the beam delivered by a straight supermirror guide. Choose the
    guide entrance width, exit width and coating so that, at the sample
    position d_sample metres after the guide exit, the beam's horizontal
    spot size and horizontal divergence both match this instance's
    targets. Both are reported as the standard deviation of the
    distribution (spot in cm, divergence in degrees).

Fixed geometry for this instance (instrument parameters you must pass
    but may NOT change):
  src_wh = 0.114613
  L_in = 1.210129
  L_guide = 15.265908
  wl = 5.37935
  dwl = 0.463611
  d_sample = 0.843826

Design parameters you control (bounds inclusive):
  w_in in [0.01, 0.09]  (baseline 0.05)
  w_out in [0.01, 0.09]  (baseline 0.03)
  m_coat in [1.0, 3.0]  (baseline 2.0)

TARGETS for this instance (the episode passes when EVERY quantity is
    within its stated tolerance):
  spot size (horizontal std) on monitor 'psd': (set at reset) +/-5%
  divergence (horizontal std) on monitor 'divmon': (set at reset) +/-5%

Evaluation runs at a fixed protocol (ncount 100000, env-controlled
    seed); iterate however you like, the graded run is the env's.
\end{verbatim}}

\section{Extended Related Work}
\label{app:related}
Beyond the works discussed in \S\ref{sec:related}: detector co-design in
differentiable simulation \citep{chung2026detector}; simulator configuration
agents \citep{ho2026siga,zhang2025mooseagent,zou2025elagente,
ding2026topolyagent,ding2026colpack}; data-driven
discovery environments \citep{moussa2026d3gym}; engineering-design benchmarks
\citep{guo2025engdesign}; iterative self-optimization \citep{li2026optbench};
budget-matched comparisons of LLMs against classical search
\citep{rodrigues2026hpo}, whose conclusion our improvement tier reproduces;
analysis reproduction \citep{faroughy2026colliderbench}; general agentic gyms
\citep{liu2025gem}; and self-training methods related to our failed supervised
runs \citep{singh2024rest,aksitov2023restreact,pan2025swegym}.

\section{Reward Failures Found During Calibration}
\label{app:redteam}
All nine failures found in the reward and task generator. The first five were
caught during calibration and shakedown; findings 6--9 changed the
environment's design and are discussed in \S\ref{sec:integrity}.

\begin{table}[h]
\caption{The nine reward failures, where each was found, and its fix.}
\label{tab:redteam}
\centering\small
\begin{tabular}{@{}rp{0.14\linewidth}p{0.40\linewidth}p{0.31\linewidth}@{}}
\toprule
\# & Where & What went wrong & Fix \\
\midrule
1 & T2 calibration & Unconstrained flux maximization collected 672\,n/s of
direct beam past the beamstop; an upper limit on beam width cannot catch this,
because leakage narrows the pattern. & Two-sided band constraints relative to
the baseline pattern. \\
2 & T2 calibration & The simulator's Nelder--Mead optimizer ignored parameter
bounds (a width escaped to 2.1\,m). & Classical baselines are bounds- and
constraint-filtered ensembles. \\
3 & T2 calibration & Winner's curse: the best of 30 noisy evaluations fell
below its own target at a fresh seed. & Re-verify any selected optimum. \\
4 & Agent shakedown & Matching monitors by event count graded an agent's
diagnostic monitor as its detector; correct instruments from two models failed
3/6. & Match monitors by role and position. \\
5 & Phase-space check & A legitimate action read $1.14\times$ the bound on a
250-event monitor; at $10\times$ the rays it read 0.99. & Evaluate the bound
only above the statistics floor. \\
6 & SANS family & Maximal pinholes, with no model, passed 8/25 held-out
instances, one at 11{,}992$\times$ the target: unscattered beam leaked around
the beamstop and counted as signal. 22 of 35 cached optima were such leaks. &
Geometric direct-beam check at L1. \\
7 & Both families & One fixed action passed about half of all held-out
instances (guide 52.3\%, SANS 51.0\%; $n=300$). & Per-instance specifications
and baselines, calibration under common random numbers, and the gate below. \\
8 & Guide family, repaired & Passed the constant-policy gate at every bar, yet
a rule that copies the two limits stated in the prompt passes 96.3\% of
held-out instances. & Report a no-model readout-rule baseline beside every
training result. \\
9 & Bender family, first version & Passed the gate at 10.7\%, yet on 24.7\% of
instances the hidden design filtered nothing, so the targets were the source
spectrum's own statistics and any transparent design solved them. & Require
hidden designs to alter the observable (cut 25--85\% of the incident band);
diagnose with design concentration. \\
\bottomrule
\end{tabular}
\end{table}

\section{The Probe Suite, Family by Family}
\label{app:probes}
Every family faces three no-model probes before its pass rates are read as
capability: the best \emph{fixed action} (a grid over the free parameters, the
baseline, and every probed instance's own solution applied to all others); the
best \emph{rule} that reads quantities stated in the prompt, including
hand-coded physics inversions; and the \emph{concentration} of a trained
policy's own passing designs, which bounds what a whole class of answers can
do rather than a single point. A family is admitted only if the one-sided 95\%
upper bound on the first stays under 20\% and no copy-type rule succeeds; a
physics inversion is reported as a reference arm, not a disqualification.

Two details decide whether the gate has any power. \emph{Candidate coverage}: a
five-level grid over each free parameter reported the SANS family's best
constant at 12\%, while the true best, lying between grid points, passed 51\%,
so candidates now come from the grid, the baseline, every probed instance's own
calibrated optimum, and a local refinement around the best of them.
\emph{Sample size}: maximizing over hundreds of candidates inflates small
samples (the guide's best-constant share at the released bar fell from 20\%
at $n=25$ to 13.3\% at $n=150$), and against the 20\% ceiling, $n=25$
certifies only at $\le 1/25$, $n=150$ at $\le 21/150$ and $n=300$ at
$\le 48/300$. A run observing 5/25, whose 95\% interval is 7--41\%, had earlier
been called a pass. For the maximization families the calibrated optimum exceeds
the baseline by a median $14.7\times$ (guide) and $10.1\times$ (SANS), so the
bar leaves real headroom.

\begin{table}[h]
\caption{Probes per family, on 150 held-out instances unless noted. Four
designs were discarded: an undersized-baseline guide version that predates the
gate (\S\ref{sec:environment}, not shown), the guide family below, where a
rule copying two limits printed in the prompt passed 96.3\% of the 300
instances used for training evaluation (98.0\% on the 150-instance probe),
the SANS collimation family, kept for evaluation only, and the first bender
family, where the gate passed but a quarter of instances had a hidden design
that filtered nothing, visible only as concentration.}
\label{tab:probes}
\centering\small
\begin{tabular}{@{}llccl@{}}
\toprule
Family & Objective & Fixed / lookup & Readout rule & Verdict \\
\midrule
Guide divergence & maximize & 13.3\% & \textbf{96.3\%} & rejected \\
SANS collimation & maximize & 14.0\% & \textbf{56.0\%} & evaluation only \\
Bender, first version & match & 10.7\% & --- & rejected (concentration) \\
\midrule
Guide match & match & \phantom{0}4.0\% & 27.7\%$^{\dagger}$ & admitted \\
SANS match & match & 12.7\% & \phantom{0}3.3\%$^{\dagger}$ & admitted \\
TOF chopper & match & \phantom{0}2.7\% & \phantom{0}0.0\% & admitted \\
Bender & match & \phantom{0}8.0\% & 12.7\%$^{\dagger}$ & admitted \\
\bottomrule
\end{tabular}

\smallskip
{\footnotesize $^{\dagger}$Physics inversion, reported as a reference arm.
Concentration of passing designs on admitted families, for the checkpoints
Figure~\ref{fig:families} reports: 225/230 distinct (guide match), 115/133
(SANS match), 138/140 (TOF chopper), 63/227 (bender, one design reused 18
times). The admitted bender is \emph{more} concentrated than the rejected
version (28\% distinct against 42/141 = 30\%): its two observables both follow
from one cutoff wavelength, so a low-dimensional solution set is intrinsic
there, and what condemned the rejected version was not concentration alone
but the quarter of instances whose hidden design filtered nothing.
Concentration is a diagnostic, not a gate: no threshold is applied.}
\end{table}

\section{Procedural Family Definitions}
\label{app:families}
Table~\ref{tab:families} defines the six families, with the parameters the
agent controls, the objective or specification it is graded on, the held-out
regime, and each family's status after the gate of \S\ref{sec:integrity}.
\begin{table}[!ht]
\caption{Procedural families and their status. Specification checks run at L1
before simulation and band constraints at L3, on the two maximization families
only; the matching families carry neither and are graded on hidden-design
targets. Tolerances were fixed before training by measuring the untrained 8B
(for SANS match the sweep read 45.3\% at $\pm3\%$, 24.7\% at $\pm1.5\%$
and 18.0\% at $\pm1\%$, and $\pm1.5\%$ was adopted); the sweeps are in the
release.}
\label{tab:families}
\centering\small
\begin{tabular}{@{}p{0.15\linewidth}p{0.24\linewidth}p{0.27\linewidth}p{0.24\linewidth}@{}}
\toprule
Family & Free parameters & Objective / specification & Held-out regime \\
\midrule
Guide (rejected) & entrance and exit width $\in[1,9]$\,cm; $m\in[1,3]$ &
maximize flux within the divergence limit $m\cdot0.099^\circ\!/\text{\AA}\cdot\lambda
\le \theta_{\max}$; exit $\le$ sample width &
guide 12.5--16\,m (train 6--12) \\
SANS (evaluation only) & two pinhole radii $\in[1,20]$\,mm &
maximize scattered intensity; beam inside the beamstop, $q_{\min}\le
1/R_{\text{sphere}}$, beam fits the sample &
collimation 4.5--6\,m (train 2.5--4) \\
Guide match (admitted) & as guide, without the divergence-limit check; sample $0.1$--$1.5$\,m after the exit &
match spot size and divergence within $\pm5\%$ &
guide 12.5--16\,m (train 6--12) \\
SANS match (admitted) & two pinhole radii $\in[1,20]$\,mm &
match beam width at the sample and at the beamstop within $\pm1.5\%$ &
collimation 4.5--6\,m (train 2.5--4) \\
TOF chopper (admitted) & chopper frequency 20--60\,Hz, phase, opening angle &
match mean wavelength and spread at the sample within $\pm5\%$ &
chopper separation 9.5--12\,m (train below) \\
Bender (admitted) & radius of curvature $40$--$550$\,m; aperture width $2$--$8$\,cm;
$m\in[1.5,4]$ &
match the transmitted band's mean wavelength within $\pm0.25\%$ and its spread
within $\pm1\%$; the hidden design must cut 25--85\% of the incident band &
bender 23--30\,m (train 12--22) \\
\bottomrule
\end{tabular}
\end{table}

\section{The Phase-Space Bound}
\label{app:bound}
The check of \S\ref{sec:environment} is derived for the two maximization
families and is inactive elsewhere: the environment's bound table has entries
for the guide and SANS collimation families only, and every other family
passes the check unconditionally. For the guide family the bound is an
\'etendue bound at the divergence monitor: $A$ is the monitor's window,
$\Omega$ the solid angle of its $\pm\theta_{\max}$ acceptance on both axes,
$\Delta\lambda$ the full source band, and $\Phi$ the template's unit
brilliance. For the SANS family it is a conservation bound through the pinhole
chain: nothing downstream can exceed what the second pinhole admits, so $A$ is
that pinhole's area, $\Omega$ the solid angle the first pinhole subtends from
it, $\Delta\lambda$ the source band and $\Phi$ the template's brilliance; the
bound moves with the agent's radii. Scattering redistributes into $4\pi$ and
the detector sees far less than the bound, which is why the check is loose
there and the band constraints do the fine-grained work. The margin of $1.10$
was calibrated on the guide family at $10^5$--$10^6$ rays: legitimate optima
read 0.93--0.99 of the bound, and a stress test of 56 corner, boundary and
random-extreme actions produced no false flag with a maximum legitimate
utilization of $1.016$, the excess over 1 coming from the source component's
weighting approximation and protocol-level statistics rather than from the
optics. Evaluating the bound only above the statistics floor is what keeps
Monte Carlo noise from flagging a correct design (Table~\ref{tab:redteam},
finding 5). Bounding the reward by the phase-space ceiling was proposed for
classical guide optimization by \citet{bertelsen2016phasespace}.

\section{Training Details}
\label{app:training}

\paragraph{States.} A training set is every turn of the untrained model's
episodes over 300 training instances of a family, failures included: about
2{,}810 states for guide match, 8{,}592 for the pooled three-family set. A state
is the full dialogue up to a decision point; the model is trained on the action
it takes there, not on a whole episode.

\paragraph{GRPO.} Each update samples 8 states and 8 completions per state at
temperature 1.0, capped at 96 new tokens per action with prompts truncated at
6{,}000 tokens, scores each completion with the reward ladder through a
simulation server, and applies a group-normalized policy gradient with a KL
penalty to the frozen base: the advantage is $(r - \bar r)/\sigma$ within a
group (groups with $\sigma < 10^{-3}$ contribute nothing), each sampled batch
is used for one gradient step on the token log-probabilities of its
completions, the KL term is the per-token $k_3$ estimator against the frozen
base, and gradients are norm-clipped at 1.0. Optimizer AdamW; seed 20260916;
thinking disabled in the chat template. Training runs in two stages: 120
updates at learning rate $10^{-5}$ with KL weight 0.02, then 100 more at
$3\times10^{-5}$ with KL 0.01, 220 updates in all, about 3.5 hours on one
A100. The pass/fail ablation and the joint runs use this same schedule, so they
differ from the headline run only in the quantity under test. The 120- and
220-update checkpoints are the stage boundaries; the 340-update point in
\S\ref{sec:analysis} continues the second stage for 120 more updates, and
that three-point sweep is scored on a second held-out slice, not the headline
slice. GPU time for the whole study was not metered; the per-run figure above
is the unit.

\paragraph{LoRA.} Rank 32, $\alpha = 64$, applied to the attention projections
(\texttt{q}, \texttt{k}, \texttt{v}, \texttt{o}) and the MLP projections
(\texttt{gate}, \texttt{up}, \texttt{down}) of Qwen3-8B. The merged adapter is
served through vLLM for evaluation, and the served model is checked against the
training tokenization before any number is read from it.

\paragraph{Imitation.} The same LoRA shape, trained on the untrained model's own
successful episodes as chat dialogues: learning rate $10^{-4}$, 2 epochs, about
32{,}000 tokens per step, sequences up to 8{,}192 tokens with the prompt masked
out of the loss, seed 20260913. A dialogue whose completion would be truncated
is dropped rather than clipped, so the model is never trained on half an answer.
Table~\ref{tab:imitation} lists the runs.

\begin{table}[h]
\caption{Supervised imitation runs, each read against its own untrained
baseline on the pre-repair task version it ran on. All four predate the
2026-09-16 feedback-label fix (\S\ref{sec:analysis}) and none is comparable
with the RL rows of Table~\ref{tab:buys}.}
\label{tab:imitation}
\centering\small
\begin{tabular}{@{}llccr@{}}
\toprule
Run & Family & Pass rate & Read against & $\Delta$ \\
\midrule
Imitation of all turns (pre-registered) & guide, pre-repair & 31.3\% & 40.3\% untrained & $-9$ \\
Imitation of passing turns & guide, pre-repair & 52.3\% & 52.3\% by one fixed action & $0$ \\
Imitation of passing turns & SANS, pre-repair & 27.0\% & 31.0\% untrained & $-4$ \\
Same, harder targets & SANS, pre-repair & \phantom{0}1.0\% & \phantom{0}8.3\% untrained & $-7$ \\
\bottomrule
\end{tabular}
\end{table}

\section{Statistical Methods}
\label{app:stats}
Comparisons on the same instances are paired, and we report the discordant
counts rather than only a $p$-value, so the reader can see how much of a
difference rests on how few episodes.

\begin{itemize}
  \item \textbf{Paired exact McNemar} \citep{mcnemar1947} for two policies
  on one instance slice: every trained-versus-untrained comparison, the
  tolerance and turn-budget stress tests, the ablation, the transfer result,
  and the two training seeds against each other (agreement is the share of
  instances on which both seeds have the same outcome). The statistic is the
  discordant pair count, and the exact binomial form is used throughout rather
  than the $\chi^2$ approximation, which is unreliable when discordant pairs are
  few. The two guide-match seeds are compared unpaired by Fisher's test because
  the question there is whether the two rates differ, not which instances.
  \item \textbf{Fisher exact} for unpaired $2\times2$ comparisons, which is how
  the loop and one-shot arms are compared on \mcbench{}: the two arms run the
  same tasks but are separate episodes, not paired trials.
  \item \textbf{Wilson intervals} for the pass-rate error bars in
  Figure~\ref{fig:families}, which stay inside $[0,1]$ at the small rates the
  untrained arms produce, where a normal interval does not.
  \item \textbf{One-sided 95\% Clopper--Pearson upper bounds}
  \citep{clopper1934} for the constant-policy gate. The gate asks whether a
  family \emph{could} admit a fixed answer passing more than 20\% of instances,
  so it must be decided on the upper end of the interval, not on the point
  estimate; Appendix~\ref{app:probes} gives the sample sizes this implies.
\end{itemize}
We report unadjusted $p$-values. The paper contains more than a dozen tests, so
only comparisons at $p < 10^{-3}$ should be read as confirmatory; the ones near
$0.01$--$0.05$ (the held-out scaffold comparison, the seed agreement, the two
seeds' rates) are reported as observations. The frontier comparison in
Appendix~\ref{app:guidematch} is unpaired, on the first 100 instances of the
slice.

\section{Seen-Tier Results per Model}
\label{app:matrix}
Table~\ref{tab:matrix} gives the full per-model breakdown behind
Table~\ref{tab:bench}: passes and, for every failed episode, the deepest
ladder level it reached.
\begin{table}[!ht]
\caption{Seen-tier results, 16 scored tasks per cell. Failures are counted by
the deepest level reached; L0 means the episode ended without a finished
instrument (loop) or without a parseable file (one-shot). Levels are read
mechanically from the grading report. Over both arms the 185 seen-tier
failures are 89 at L0, 73 at L1, 8 at L2, 8 at L3 and 7 at L4.}
\label{tab:matrix}
\centering\small
\begin{tabular}{@{}lcc|ccccc|ccccc@{}}
\toprule
& \multicolumn{2}{c|}{Passes} & \multicolumn{5}{c|}{Loop failures} &
\multicolumn{5}{c}{One-shot failures} \\
Model & Loop & 1-shot & L4 & L3 & L2 & L1 & L0 & L4 & L3 & L2 & L1 & L0 \\
\midrule
Claude Sonnet 5      & 5 & 7 & 1 & 0 & 0 & 9 & 1  & 0 & 1 & 2 & 6  & 0 \\
Claude Haiku 4.5     & 4 & 4 & 4 & 2 & 2 & 4 & 0  & 0 & 0 & 0 & 12 & 0 \\
Gemini 3.6 Flash     & 4 & 6 & 0 & 1 & 0 & 1 & 10 & 1 & 1 & 2 & 6  & 0 \\
Gemini 3.5 Flash-Lite& 2 & 1 & 1 & 0 & 1 & 5 & 7  & 0 & 0 & 0 & 15 & 0 \\
Llama 4 Maverick     & 0 & 2 & 0 & 2 & 0 & 1 & 13 & 0 & 0 & 0 & 14 & 0 \\
Qwen3-32B            & 3 & 0 & 0 & 0 & 0 & 0 & 13 & 0 & 0 & 0 & 0  & 16 \\
Qwen3-8B             & 1 & 0 & 0 & 1 & 1 & 0 & 13 & 0 & 0 & 0 & 0  & 16 \\
\bottomrule
\end{tabular}
\end{table}

\section{Joint Multi-Family Training}
\label{app:joint}
Joint training pools the recorded decision points of several families into one
training set and trains a single LoRA on the mixture, starting from the same
base model the specialists start from; no specialist checkpoint is ever
warm-started. The budget is one specialist's, 220 updates, split across the
families, so one policy and one training run stand against four policies and
four training runs.

\begin{table}[h]
\caption{Joint runs, all evaluated on held-out instances 300--599 at 10 turns
and temperature 0. \texttt{joint3rep} repeats \texttt{joint3} exactly (same
families, same 8{,}592 states, same budget and hyperparameters) with only the
seed changed, and lands at the untrained rate on TOF chopper. Its stage-A
checkpoint was already behind (11.0\% against \texttt{joint3}'s eventual
55.0\%), so the two runs diverged early. Only this three-family configuration
was ever run twice: there is no evidence that four-family pooling is the stable
one. $^{*}$The SANS-match specialist's 51.0\% is read on the slice its
checkpoint was selected on; on a slice chosen for nothing the same checkpoint
gives 44.3\%, the number Figure~\ref{fig:families} reports; the untrained-8B
row is on the 300--599 slice, where SANS match reads 22.3\%, against 26.0\% on
the selection-free slice Figure~\ref{fig:families} draws, so the 18-point
SANS-match gain quoted in \S\ref{sec:training} is the figure's
($44.3 - 26.0$), not this table's. The other three specialist cells are
like-for-like with the joint runs.}
\label{tab:joint}
\centering\small
\begin{tabular}{@{}llcccc@{}}
\toprule
Run & Pooled & Guide match & SANS match & TOF chopper & Bender \\
\midrule
Specialists, one per family & --- & 76.7\% & 51.0\%$^{*}$ & 46.7\% & 75.7\% \\
\addlinespace
\texttt{joint3}    & 3 families & 67.7\% & 24.0\% & 55.0\% & --- \\
\texttt{joint4}    & 4 families & 63.7\% & 43.3\% & 56.3\% & 55.0\% \\
\texttt{joint3rep} & 3, new seed & 34.0\% & 19.3\% & \phantom{0}2.3\% & --- \\
\addlinespace
\texttt{joint3rep}, stage A & & --- & --- & 11.0\% & --- \\
Untrained 8B & --- & 11.3\% & 22.3\% & \phantom{0}2.0\% & 23.0\% \\
\bottomrule
\end{tabular}
\end{table}

The failure is quiet rather than loud: mean reward stayed flat near 0.77--0.83
throughout, and only the held-out evaluation revealed it. What does show it
during training is the share of sampled groups carrying any gradient (groups
whose completions all score alike contribute no advantage), which held near
4.5 of 8 in \texttt{joint3} and fell to 2.1 of 8 over the second stage of
\texttt{joint3rep}.

\section{Guide Match: All Policies and Baselines}
\label{app:guidematch}
Table~\ref{tab:training} lists every arm run on the guide-match headline
slice, from the no-model baselines through the trained policies to the
frontier models, on which \S\ref{sec:analysis} draws.
\begin{table}[!ht]
\caption{Guide match, held-out instances 300--599, a slice used for nothing but
this measurement. Agents get 10 turns at temperature 0; classical optimizers get
the same simulation budget and act on the parametrization directly, which the
agent reads from its task sheet and reaches only through tool calls. Italic
rows are no-model baselines. Frontier models were run on the first 100
instances of the slice, where the trained model scores 76 of 100.}
\label{tab:training}
\centering\small
\begin{tabular}{@{}lcl@{}}
\toprule
Policy & Passed & Note \\
\midrule
Untrained Qwen3-8B & 34/300 \phantom{0}(11.3\%) & exhausts the 10-turn cap \\
Untrained Qwen3-32B & 34/300 \phantom{0}(11.3\%) & scale buys nothing here \\
\emph{Best fixed design or lookup} & \phantom{00/300 }\emph{(4.0\%)} & incl.\ other instances' solutions \\
\emph{Nelder--Mead, 10 simulations} & \emph{8/300} \phantom{00}\emph{(2.7\%)} & 40.0\% at 30 simulations \\
\emph{Random search, 10 simulations} & \emph{22/300} \phantom{0}\emph{(7.3\%)} & 20.0\% at 30 \\
\emph{Coordinate descent, 10 simulations} & \emph{40/300} \emph{(13.3\%)} & 73.7\% at 30 \\
\emph{Coordinate descent, random start, 10 sims} & \emph{43/300} \emph{(14.3\%)} & vs.\ baseline start 34/31, $p = 0.80$ \\
\emph{One-shot physics formula} & \emph{83/300} \emph{(27.7\%)} & optics inversion, best of 36 \\
\emph{Formula, then coordinate descent, 10 sims} & \emph{243/300} \emph{(81.0\%)} & 59 vs.\ 46 paired against GRPO, $p = 0.24$ \\
\textbf{GRPO Qwen3-8B (seed 1)} & \textbf{230/300 (76.7\%)} & 213 vs.\ 17 paired against the 8B \\
\textbf{GRPO Qwen3-8B (seed 2)} & \textbf{207/300 (69.0\%)} & independent repeat \\
GRPO, pass/fail reward only & 50/300 \phantom{0}(16.7\%) & ladder ablation \\
\midrule
Claude Sonnet 5 & 99/100 \phantom{0}(99\%) & \multirow{2}{*}{\$9.69 for both} \\
Gemini 3.6 Flash & 98/100 \phantom{0}(98\%) & \\
\bottomrule
\end{tabular}
\end{table}

\end{document}